\documentclass[11pt]{article}
\usepackage{coling2020}
\usepackage{times}
\usepackage{url}
\usepackage{latexsym}
\usepackage{xcolor}

\colingfinalcopy 

\title{UPB at SemEval-2020 Task 12: Multilingual Offensive Language Detection on Social Media by Fine-tuning a Variety of BERT-based Models}

\author{Mircea-Adrian Tanase, Dumitru-Clementin Cercel, Costin-Gabriel Chiru \\
  University Politehnica of Bucharest, Faculty of Automatic Control and Computers \\
    \tt {mircea.tanase@stud.acs.upb.ro} \\
    {\tt \{dumitru.cercel, costin.chiru\}@cs.pub.ro} \\
  }
  
\date{}

\begin{document}
\maketitle
\begin{abstract}
  Offensive language detection is one of the most challenging problem in the natural language processing field, being imposed by the rising presence of this phenomenon in online social media. This paper describes our Transformer-based solutions for identifying offensive language on Twitter in five languages (i.e., English, Arabic, Danish, Greek, and Turkish), which was employed in Subtask A of the Offenseval 2020 shared task. Several neural architectures (i.e., BERT, mBERT, Roberta, XLM-Roberta, and ALBERT), pre-trained using both single-language and multilingual corpora, were fine-tuned and compared using multiple combinations of datasets.
  Finally, the highest-scoring models were used for our submissions in the competition, which ranked our team 21st of 85, 28th of 53, 19th of 39, 16th of 37, and 10th of 46 for English, Arabic, Danish, Greek, and Turkish, respectively.
\end{abstract}

\section{Introduction}
\label{intro}

\blfootnote{This work is licensed under a Creative Commons Attribution 4.0 International Licence. Licence details: http://creativecommons.org/licenses/by/4.0/.
}

Social media platforms are gaining increasing popularity for both personal and political communication. Recent studies uncovered disturbing trends in communications on the Internet. For example, Pew Research Center\footnote{\url{https://www.pewresearch.org/internet/2017/07/11/online-harassment-2017/}} discovered that 60\% of Internet users have witnessed a form of online harassment, while 41\% of them have personally experienced it. The majority of the latter category says the most recent such experience occurred on a social media platform. Although most of these platforms provide ways of reporting offensive or hateful content, only 9\% of the victims have considered using these tools.

Traditionally, identifying and removing offensive or hateful content on the Internet is performed by human moderators that inspect each piece of content flagged by the users and label it appropriately. This process has two major disadvantages. As previously mentioned, the first one is that the proportion of users that even considered using the tools provided by the platforms in order to report the harmful content is very small. The second one is represented by the continuously growing volume of data that needs to be analyzed. 

However, the task of automated offensive language detection on social media is a very complicated problem. This is because the process of labeling an offensive language dataset has proved to be very challenging, as every individual reacts differently to the same content, and the consensus on assigning a label to a piece of content is often difficult to obtain \cite{waseem2017understanding}.


The SemEval-2019 shared task 6~\cite{zampieri2019semeval}, \textit{Offenseval 2019}, was the first competition oriented towards detecting offensive language in social media, specifically Twitter. The SemEval-2020 shared task 12~\cite{zampieri-etal-2020-semeval}, \textit{Offenseval 2020}, proposes the same problem, the novelty being a very large automatically labeled English dataset and also smaller datasets for four other languages: Arabic, Danish, Greek, and Turkish. This paper describes the Transformer-based machine learning models we used in our submissions for each language in Subtask A, where the goal is to identify the offensive language in tweets – a binary classification problem.

The remainder of the paper is structured as follows: in section~\ref{relwork}, a brief analysis of the state-of-the-art approaches is performed. In section~\ref{methodology}, the methods employed for automated offensive language detection are presented. Section~\ref{sec:data} describes the data used in this study. Section~\ref{sec:evaluation} presents the evaluation process, and in section~\ref{sec:conclusions}, conclusions are drawn.

\section{Related work}
\label{relwork}

Automating the task of offensive language detection becomes a necessity on the Internet of today, especially on communication platforms. The efforts in this direction have substantially grown in the research community. The first approaches on a related subject include the detection of racist texts in web pages~\cite{greevy2004classifying}, where the authors considered part-of-speech tags as inputs for support vector machines.

This type of problems gained a lot of interest in the last decade, as advanced machine learning techniques has been developed for NLP tasks, and also computing power became increasingly accessible. Cambria et.al.~\shortcite{cambria2010not} proposed the detection of web trolling (i.e., posting outrageous messages that are meant to provoke an emotional response). Hate speech and offensive language detection in Twitter samples was  analyzed by Davidson et al.~\shortcite{davidson2017automated}. Their study presents a framework for differentiating between profanity and hate speech and also describes the annotation process of such a dataset. Moreover, their experiments with various text preprocessing techniques are described and the logistic regression algorithm is used for hate speech and offensive language classification.
Malmasi  and Zampieri~\shortcite{malmasi2018challenges} continued the experiments on this dataset using n-gram and skip-gram features.

More recently, neural networks gained interest in this type of problems. For example, Gamb\"{a}ck and Sikdar~\shortcite{gamback2017using} relied on a Convolutional Neural Network (CNN) \cite{kim2014convolutional}  to surpass
the state-of-the-art results on the previously mentioned dataset.
~\newcite{zhang2018detecting} also improved these results by combining two different deep learning architectures, namely CNN and Gated Recurrent Unit \cite{cho2014learning}.

There are a series of surveys and shared-tasks that took place in the last years on the subject of detecting the online offensive, abusive, hateful, or toxic content. 
Schmidt and Wiegand~\shortcite{schmidt2017survey} introduced a comprehensive survey on different methods to automatically recognize hate speech, focusing mostly on non-neural network approaches. Shared-tasks analyzing problems in the same areas include both editions of Abusive Language Online~\cite{fivser2018proceedings}, which focused mostly on cyberbullying, TRAC~\cite{kumar2018benchmarking}, which mainly studied aggressiveness, HASOC~\cite{mandl2019overview}, which also addressed the problem of hate speech, the same as the SemEval-2019 Task 5~\cite{basile2019semeval} competition.

\section{Methodology}
\label{methodology}

\subsection{Baseline}
\label{ssec:baseline}

As a baseline, we used a non-neural network approach, which employs the XGBoost~\cite{chen2016xgboost} algorithm for classification and multiple text processing techniques for feature extraction:

\begin{itemize}
	\item Firstly, the lemma of the words was extracted and the TF-IDF scores were computed for the n-grams obtained, with n = 1, 2, 3.
	\item Secondly, part-of-speech tags were extracted using the NLTK Python package~\cite{loper2002nltk} and the TF-IDF scores were computed for the tag n-grams obtained, with n = 1, 2, 3.
	\item Thirdly, TF-IDF scores were computed for character n-grams, with n = 1, 2, 3.
	\item Sentiment analysis features were obtained using the VADER tool~\cite{hutto2015computationally}, which is based on a mapping between lexical features and sentiment scores.
	\item Finally, other lexical features were added, such as the number of characters, words, syllables, and the Flesch-Kincaid readability score~\cite{kincaid1975derivation}.
\end{itemize}

\subsection{BERT}
\label{ssec:BERT}

Bidirectional Encoder Representations from Transformers (BERT)  ~\cite{devlin2018bert} is a novel deep learning architecture designed for NLP tasks by the Google team. It combines the WordPiece embeddings~\cite{wu2016google} and the Transformers ~\cite{wolf2019transformers} which represents the last major breakthrough in NLP. The BERT models significantly outperformed the state-of-the-art approaches on various text classification or question answering benchmarks.
The architecture is a multi-layer transformer encoder, with the novelty consisting of the usage of bidirectional attention instead of recurrent units.

The BERT model can be used for any NLP classification task using a technique called fine-tuning. This consists of starting with a model that has been pre-trained on a very large and comprehensive dataset and training it further for the respective classification task, in our case offensive language identification. There are several pre-trained BERT versions available, differing in terms of model size (i.e., the number of transformers) and the corpora used for pre-training. Therefore, we experimented  with the following BERT-aware models:
\begin{itemize}
	\item BERT-base, which is pre-trained on the English Wikipedia Corpus.
	\item BERT-base for Danish\footnote{\url{https://github.com/botxo/nordic_bert}}, which is  pre-trained on the entire dump of Danish Wikipedia pages.
	\item multilingual BERT (mBERT) \cite{pires2019multilingual}, which is pre-trained on a corpus containing the top 104 languages, considering the  Wikipedia pages for each language.
\end{itemize}

\subsection{Roberta and XLM-Roberta}
\label{ssec:roberta}
\newcite{liu2019roberta} analyzed the BERT model and concluded that it is under-trained, claiming that the hyperparameter choice can significantly impact the obtained results. The robust pre-training method they proposed, namely Roberta, achieved better performances in the same NLP tasks. Based on their work, XLM-Roberta was  developed by ~\newcite{conneau2019unsupervised} for multilingual NLP tasks. 
It is pre-trained  for more than 100 languages, similarly to mBERT, which it manages to outperform. An interesting observation is that the more significant improvement in results was obtained for low-represented languages, which recommends using it on all five languages in the subtask A of the current competition.
Here, we used the base architectures of Roberta for English and XLM-Roberta, both pre-trained using large amounts of specific CommonCrawl\footnote{\url{https://commoncrawl.org/}} data.

\subsection{ALBERT}
\label{ssec:albert}

ALBERT \cite{lan2019albert},  A lite BERT, is a BERT variation that brings two novel parameter-reduction techniques, resulting in lower resource consumption at training time, and at the same time obtains similar performances with the original BERT model. Moreover, ALBERT uses a self-supervised loss focusing on improving modeling on inter-sentence coherence, which helps to obtain better results on NLP tasks with multi-sentence inputs. We fine-tuned the ALBERT-base model for our English language experiments.

\section{Data}
\label{sec:data}

To analyze the influence of extending the competition-provided training datasets with other corpora constructed for related tasks, we used two additional datasets for fine-tuning the previously mentioned models. The summary of each dataset is presented in Table~\ref{datasets-table}. We can observe that all the datasets have a similar structure (i.e., binary labels, unbalanced, and positive label ratio of 10\%-50\%).

\begin{table}[h]
\begin{center}
\begin{tabular}{|l|r|r|r|r|}
\hline \bf Dataset & \bf No. Samples & \bf Positive & \bf Train Set ID & \bf Validation \\
\bf  & \bf  & \bf Ratio (\%) & \bf & \bf Set ID\\ \hline
Offenseval 2020 English & 9,075,418 & 12.58 & Off\_en\_train & Off\_en\_val \\ \hline
Offenseval 2020 Arabic & 7,000 & 19.58 & Off\_ar\_train & Off\_ar\_val \\ \hline
Offenseval 2020 Danish & 3,000 & 12.80 & Off\_da\_train & Off\_da\_val \\ \hline
Offenseval 2020 Greek & 8,743 & 28.43 & Off\_gr\_train & Off\_gr\_val \\ \hline
Offenseval 2020 Turkish & 31,277 & 19.33 & Off\_tr\_train & Off\_tr\_val \\ \hline
Offenseval 2020 all lang. & 9,125,438 & 12.63 & Off\_all\_train & Off\_all\_val \\ \hline
Offenseval 2020 all lang., & & & & \\
except English & 50,020 & 20.56 & Off\_no\_eng\_train & Off\_no\_eng\_val \\ \hline
OLID & 13,240 & 33.23 & OLID & -\\ \hline
HASOC English & 5,852 & 38.63 & HASOC\_en & -\\ \hline
HASOC German & 3,819 & 10.65 & HASOC\_gr & -\\ \hline
HASOC Hindi & 4,665 & 47.07 & HASOC\_hi & -\\ \hline
HASOC all languages & 14,336 & 29.41 & HASOC\_all & -\\ \hline
\end{tabular}
\end{center}
\caption{\label{datasets-table} Statistics of datasets. }
\end{table}

\subsection{Offenseval 2020 Dataset}
\label{ssec:offenseval2020dataset}

The Offenseval 2020 dataset is composed of five subsets of tweets, one for each of the languages: English~\cite{rosenthal2020}, Arabic~\cite{mubarak2020arabic}, Danish~\cite{sigurbergsson2020offensive}, Greek~\cite{pitenis2020}, and Turkish~\cite{coltekikin2020}. The last four are similar in structure: for each sample, the Subtask A label is binary, revealing whether the tweet contains offensive language or not. 
The English subset has a more complex annotation scheme for the Subtask A: for each sample, both the average and the standard deviation of the scores assigned by a pool of semi-supervised learning models are given. After exploring the distribution of these values, the following heuristic was used for converting them to binary labels:
\begin{itemize}
	\item If the model average score is greater than 0.6, the sample is labeled as positive. 
	\item If the model average score is between 0.5 and 0.6, and the standard deviation is smaller than 0.1 (there is a consensus), the sample is labeled as positive.
	\item All other samples are labeled as negative.
\end{itemize}

Using this method, we obtained 114,223 English tweet samples labeled as positive. Finally, a 10\% ratio of each language subset was put aside for validation purposes, preserving the label distribution.

\subsection{OLID Dataset}
\label{ssec:olid}

~\newcite{zampieri2019predicting} introduced the Offensive Language Identification Dataset (OLID)  to the Offenseval 2019 shared task. Note that this was actually the starting point for the Offenseval 2020 dataset. Moreover, OLID contains only English language tweets, and the label for the Subtask A (i.e., offensive language identification) is binary.

\subsection{HASOC Dataset}
\label{ssec:hasoc}

The Hate Speech and Offensive Communication dataset\footnote{\url{https://hasocfire.github.io/hasoc/2019/dataset.html}} was proposed for the HASOC 2019 competition \cite{mandl2019overview} with the goal of identifying both hate speech and offensive content in Indo-European languages (i.e., English, German, and Hindi). Task 1 of this competition required identifying hateful or offensive tweet samples, and the labels were binary, so the dataset can be considered similar to the Offenseval 2020 subsets.

\subsection{Preprocessing}
\label{ssec:preprocessing}

The main preprocessing step is done using the BERT-specific tokenizer, which splits a sentence into tokens in a WordPiece manner. Two more Twitter-specific prior steps were performed:

\begin{itemize}
	\item Replacing the emojis with the corresponding textual representation by using the emojiPython package\footnote{\url{https://pypi.org/project/emoji/}}. 
	\item Normalizing the hashtags  (e.g., "\#MakeAmericaGreatAgain" is split into "Make", "America", "Great", and "Again").
\end{itemize}

\noindent For the non-English languages, we also explored the approach of translating the texts to English as a preprocessing step, in order to allow the usage of an English-only pre-trained model. We used the Yandex translation service\footnote{\url{https://tech.yandex.com/translate/}}, but the translation quality proved to be poor. That is, while most of the words were correctly translated, the syntax and meaning of the sentences were lost. For instance, the Turkish tweet "yeniden doğup gelsem çocuk kalır büyümezdiiiim" is translated as "re-born child grow I do I remain", while a more accurate translation is "If I were born again, I would be a child and I would not grow up".

\section{Experiments}
\label{sec:evaluation}

All the experiments were performed using an Ubuntu machine with 64GB RAM and one NVIDIA Titan X GPU. The hardware limitations are the reason we only experimented with the base versions of the previously mentioned architectures. The Transformer Python package~\cite{wolf2019transformers} was used for training and evaluating the Transformer-based models, and each model was fine-tuned for four epochs. 
We also used the Adam algorithm \cite{kingma2014adam} with weight decay for optimization and a learning rate of  \textit{2e-5}. 

For each language, multiple experiments were performed with the same architecture using several combinations of datasets for fine-tuning, in order to assess the impact each of them has on the performance of a certain model. This implies that, for some experiments, several datasets were simply concatenated and used as a single fine-tuning set. No additional handling of the data was required, as all of the data shared the binary label structure presented in Section~\ref{sec:data}.

The results we obtained on the validation datasets are summarized in Tables ~\ref{english-res-table},~\ref{arabic-res-table},~\ref{danish-res-table},~\ref{greek-res-table}, and~\ref{turkish-res-table} for English, Arabic, Danish, Greek, and Turkish, respectively. The reported metrics are computed for the Offenseval 2020 language-specific validation sets as described in Section~\ref{sec:data}. For each language, the highest validation set F1-score is highlighted, meaning that the corresponding model was selected and employed for predicting the language-specific competition test data in our final submission. 

\begin{table}[h]
\begin{center}
\begin{tabular}{|l|r|r|r|r|r|r|r|}
\hline

\bf Model & \bf Pre-train & \bf Preprocessing & \bf Fine-Tuning & \bf Acc & \bf Pr  & \bf Rec  & \bf F1 \\ 
\bf Architecture & \bf Language & \bf Particularities & \bf Dataset & \bf (\%) & \bf (\%)  & \bf (\%)  & \bf(\%) \\ 
\hline

Baseline & - & n-gram TFIDF & Off\_en\_train & 86.63 & 81.19 & 76.96 & 79.01\\
 &  & POS tags &  &  &  &  &
\\ \hline

BERT & English & - & Off\_en\_train & 97.90 & 91.70 & 90.69 & 91.19
\\ \hline

BERT & English & - & Off\_en\_train+ & 94.62 & 89.91 & 87.75 & 88.81 \\ 
 &  &  & OLID+HASOC\_en & & & & \\ 
\hline

mBERT & Multi & - & Off\_all\_train+ & 97.54 & 91.42 & 89.92 & 90.66 \\ 
 &  &  & OLID+HASOC\_all & & & & \\ 
\hline

Roberta & English & - &  Off\_en\_train & 98.05 & 93.11 & 91.43 & \bf 92.26 \\
\hline

Roberta & English & - & Off\_en\_train+ & 98.01 & 93.03 & 91.45 & 92.23 \\ 
 &  &  & OLID+HASOC\_en & & & & \\ 
\hline

ALBERT & English & - & Off\_en\_train+ & 97.96 & 92.10 & 91.83 & 91.96 \\
\hline

ALBERT & English & - & Off\_en\_train+ & 97.79 & 94.04 & 88.13 & 90.99 \\
 &  &  & OLID+HASOC\_en & & & & \\ 
\hline

XLM-Roberta & Multi & - & Off\_all\_train+ & 95.27 & 90.10 & 87.84 & 88.95 \\ 
 &  &  & OLID+HASOC\_all & & & & \\ 
\hline

\end{tabular}
\end{center}
\caption{\label{english-res-table} Results obtained for English (on the Off\_en\_val validation subset). }
\end{table}

\begin{table}[h]
\begin{center}
\begin{tabular}{|l|r|r|r|r|r|r|r|}
\hline

\bf Model & \bf Pre-train & \bf Preprocessing & \bf Fine-Tuning & \bf Acc & \bf Pr  & \bf Rec  & \bf F1 \\ 
\bf Architecture & \bf Language & \bf Particularities & \bf Dataset & \bf (\%) & \bf (\%)  & \bf (\%)  & \bf(\%) \\ 
\hline

Baseline & - & n-gram TFIDF & Off\_ar\_train & 56.58 & 28.52 & 22.04 & 24.86
\\ \hline

BERT&English & Translation & Off\_en\_train & 78.18 & 60.52 & 57.86 & 58.96
\\ \hline

mBERT & Multi & - & Off\_ar\_train & 81.14 & 65.42 & 64.38 & 64.89 \\ 
\hline

mBERT & Multi & - & Off\_no\_eng\_train & 86.12 & 70.48 & 68.16 & 69.30 \\
\hline

mBERT & Multi & - & Off\_all\_train & 89.90 & 72.63 & 71.88 & 72.25 \\
\hline

mBERT & Multi & - & Off\_all\_train+ & 89.44 & 74.81 & 71.66 & 73.20 \\ 
 &  &  & OLID+HASOC\_all & & & & \\ 
\hline

XLM-Roberta & Multi & - & Off\_ar\_train & 84.42 & 68.14 & 66.78 & 67.45 \\ 
\hline

XLM-Roberta & Multi & - & Off\_no\_eng\_train & 89.28 & 72.32 & 74.19 & 73.24 \\
\hline

XLM-Roberta & Multi & - & Off\_all\_train & 90.05 & 76.92 & 70.06 & 73.82 \\
\hline

XLM-Roberta & Multi & - & Off\_all\_train+ & 90.56 & 76.82 & 74.83 & \bf 75.81 \\ 
 &  &  & OLID+HASOC\_all & & & & \\ 
\hline

\end{tabular}
\end{center}
\caption{\label{arabic-res-table} Results obtained for Arabic (on the Off\_ar\_val validation subset). }
\end{table}

\begin{table}[h]
\begin{center}
\begin{tabular}{|l|r|r|r|r|r|r|r|}
\hline

\bf Model & \bf Pre-train & \bf Preprocessing & \bf Fine-Tuning & \bf Acc & \bf Pr  & \bf Rec  & \bf F1 \\ 
\bf Architecture & \bf Language & \bf Particularities & \bf Dataset & \bf (\%) & \bf (\%)  & \bf (\%)  & \bf(\%) \\ 
\hline

Baseline & - & n-gram TFIDF & Off\_da\_train & 56.92 & 27.36 & 22.39 & 24.62
\\ \hline

BERT & English & Translation & Off\_en\_train & 82.62 & 57.12 & 42.09 & 46.41
\\ \hline

mBERT & Danish & - & Off\_da\_train & 93.58 & 77.12 & 65.29 & \bf 70.71
\\ \hline

mBERT & Multi & - & Off\_da\_train & 89.76 & 64.93 & 50.12 & 56.57 \\ 
\hline

mBERT & Multi & - & Off\_no\_eng\_train & 85.36 & 66.18 & 54.68 & 59.88 \\
\hline

mBERT & Multi & - & Off\_all\_train & 90.44 & 71.06 & 60.52 & 65.36 \\
\hline

mBERT & Multi & - & Off\_all\_train+ & 90.12 & 72.41 & 58.18 & 64.51 \\ 
 &  &  & OLID+HASOC\_all & & & & \\ 
\hline

XLM-Roberta & Multi & - & Off\_da\_train & 85.42 & 66.19 & 52.94 & 58.82 \\ 
\hline

XLM-Roberta & Multi & - & Off\_no\_eng\_train & 88.14 & 68.41 & 56.30 & 61.76 \\
\hline

XLM-Roberta & Multi & - & Off\_all\_train & 91.89 & 71.87 & 60.52 & 65.71 \\
\hline

XLM-Roberta & Multi & - & Off\_all\_train+ & 91.89 & 66.66 & 73.68 & 70.00 \\ 
 &  &  & OLID+HASOC\_all & & & & \\ 
\hline

\end{tabular}
\end{center}
\caption{\label{danish-res-table} Results obtained for Danish (on the Off\_da\_val validation subset). }
\end{table}

\begin{table}[h]
\begin{center}
\begin{tabular}{|l|r|r|r|r|r|r|r|}
\hline

\bf Model & \bf Pre-train & \bf Preprocessing & \bf Fine-Tuning & \bf Acc & \bf P  & \bf R  & \bf F1 \\ 
\bf Architecture & \bf Language & \bf Particularities & \bf Dataset & \bf (\%) & \bf (\%)  & \bf (\%)  & \bf(\%) \\ 
\hline

Baseline & - & n-gram TFIDF & Off\_gr\_train & 58.42 & 26.37 & 22.28 & 24.15
\\ \hline

BERT & English & Translation & Off\_en\_train & 68.48 & 62.53 & 52.64 & 57.16
\\ \hline

mBERT & Multi & - & Off\_gr\_train & 80.76 & 71.62 & 68.31 & 69.92 \\ 
\hline

mBERT & Multi & - & Off\_no\_eng\_train & 81.14 & 70.85 & 67.18 & 68.96 \\
\hline

mBERT & Multi & - & Off\_all\_train & 82.50 & 73.80 & 61.83 & 67.28 \\
\hline

mBERT & Multi & - & Off\_all\_train+ & 84.12 & 72.24 & 67.14 & 69.59 \\ 
 &  &  & OLID+HASOC\_all & & & & \\ 
\hline

XLM-Roberta & Multi & - & Off\_gr\_train & 79.14 & 68.23 & 65.98 & 67.08 \\ 
\hline

XLM-Roberta & Multi & - & Off\_no\_eng\_train & 82.57 & 71.04 & 68.67 & 69.83 \\
\hline

XLM-Roberta & Multi & - & Off\_all\_train & 82.97 & 73.80 & 62.24 & 67.53 \\
\hline

XLM-Roberta & Multi & - & Off\_all\_train+ & 83.54 & 71.96 & 69.07 & \bf 70.49 \\ 
 &  &  & OLID+HASOC\_all & & & & \\ 
\hline

\end{tabular}
\end{center}
\caption{\label{greek-res-table} Results obtained for Greek (on the Off\_gr\_val validation subset). }
\end{table}

\begin{table}[h]
\begin{center}
\begin{tabular}{|l|r|r|r|r|r|r|r|}
\hline

\bf Model & \bf Pre-train & \bf Preprocessing & \bf Fine-Tuning & \bf Acc & \bf P  & \bf R  & \bf F1 \\ 
\bf Architecture & \bf Language & \bf Particularities & \bf Dataset & \bf (\%) & \bf (\%)  & \bf (\%)  & \bf(\%) \\ 
\hline

Baseline & - & n-gram TFIDF & Off\_tr\_train & 61.18 & 28.83 & 19.02 & 22.91
\\ \hline

BERT & English & Translation & Off\_en\_train & 85.04 & 72.63 & 36.32 & 48.42
\\ \hline

mBERT & Multi & - & Off\_tr\_train & 83.90 & 63.12 & 54.63 & 58.56 \\ 
\hline

mBERT & Multi & - & Off\_no\_eng\_train & 80.62 & 62.18 & 52.13 & 57.87 \\
\hline

mBERT & Multi & - & Off\_all\_train & 83.92 & 65.26 & 55.13 & 59.76 \\
\hline

mBERT & Multi & - & Off\_all\_train+ & 84.37 & 66.19 & 57.12 & 61.32 \\ 
 &  &  & OLID+HASOC\_all & & & & \\ 
\hline

XLM-Roberta & Multi & - & Off\_tr\_train & 81.14 & 62.44 & 52.68 & 57.14 \\ 
\hline

XLM-Roberta & Multi & - & Off\_no\_eng\_train & 86.28 & 68.25 & 54.38 & 60.53 \\
\hline

XLM-Roberta & Multi & - & Off\_all\_train & 86.82 & 69.73 & 56.36 & \bf 62.34 \\
\hline

XLM-Roberta & Multi & - & Off\_all\_train+ & 86.22 & 67.61 & 55.20 & 60.78 \\ 
 &  &  & OLID+HASOC\_all & & & & \\ 
\hline

\end{tabular}
\end{center}
\caption{\label{turkish-res-table} Results obtained for Turkish (on the Off\_tr\_val validation subset). }
\end{table}

\textbf{Results on the English Subset.} 
Firstly, we observe that, although the baseline classifier does not obtain a negligible result, the Transformer-based models outperform it, even when pre-trained for multilingual tasks, proving the performance improvement that this type of models brings to the task of automated offensive language detection.

Secondly, ALBERT and Roberta perform better than the  BERT-base architecture, thus confirming their better exploitation of the Transformer’s representative power. Furthermore, we note that even the multilingual pre-trained model performs better than the English-specific baseline, although significantly worse than the English-only pre-trained models.

Furthermore, there is no evidence that adding the OLID and  HASOC English datasets to the fine-tuning data affect the results in any way, most likely because the size of these datasets is very small in comparison to the size of the Offenseval 2020 English dataset. Finally, the best performing model is Roberta fine-tuned without adding the two additional subsets.

\textbf{Results on the non-English Subsets.} 
The very low scores obtained by the baseline approach for the non-English subsets are explained by the fact that most of the pre-processing employed is English language-specific and could not be applied for other languages.
The approach of automatically translating the texts and then applying an English-language pre-trained and fine-tuned model also seems to fail, with most of the obtained F1-scores being at least 10 points lower than the best score.

Another interesting observation is that the results are constantly improving when adding more data to the fine-tuning dataset, even if the added data is in a different language than the validation set, thus proving that the multilingual model is able to learn cross-lingual features. As expected, the XLM-Roberta model outperforms mBERT in most experimental setups. As opposed to the English language results, adding the HASOC subsets seems to improve the scores, with the sole exception of the Turkish subset. This could be partially explained by the fact that the HASOC dataset contains two other languages. For instance, the German subset from the HASOC data may have brought a performance boost to the multilingual models on the Danish validation set because Danish is more related to German, being considered a North Germanic language.

Finally, an interesting particularity can be observed for the Danish dataset. The Danish language pre-trained BERT model, fine-tuned using only the very small Danish training set, outperformed even the multilingual model, fine-tuned using all the available data. This proves that, for low-represented languages, a language-specific pre-trained model performs better than a multilingual one, even with smaller amounts of fine-tuning data.

\textbf{Results on the Leaderboard.} 
The results and the rankings obtained by our submissions can be observed in Table~\ref{leaderboard-table}, in comparison to the best performing teams. The F1-scores obtained on the Offenseval 2020 Subtask A competition test sets are as follows: 91.05\%, 82.19\%, 73.80\%, 81.40\%, and 77.89\% for English, Arabic, Danish, Greek, and Turkish, respectively.

\begin{table}[h]
\begin{center}
\begin{tabular}{|l|r|r|r|r|r|r|r|}
\hline

\bf Language & \bf F1-score (\%) & \bf Our Ranking & \bf No. Participants & \bf Leader F1-score (\%) \\ 
\hline

English & 91.05 & 21 & 85 & 92.22
\\ \hline

Arabic & 82.19 & 28 & 53 & 90.17 \\ \hline
Danish & 73.80 & 19 & 39 & 81.20 \\ \hline
Greek & 81.40 & 16 & 37 & 85.20 \\ \hline
Turkish & 77.89 & 10 & 46 & 82.57 \\ \hline

\end{tabular}
\end{center}
\caption{\label{leaderboard-table} The results of our submissions on the competition test sets.}
\end{table}

\section{Conclusions}
\label{sec:conclusions}

This work presented our approaches to automatically detect offensive language in multilingual tweets, as part of SemEval-2020 Task 12. 
We proved that the deep learning solution of fine-tuning pre-trained Transformer-based models can be used successfully to classify offensive language, and we experimented with several such architectures, fine-tuned on multiple combinations of datasets. 
Comparing the validation set performances against the test set results, we discovered that the last of them were better for the non-English languages, which shows that our models generalize well and also that the proposed test data may be easier to classify than the development data.
The smallest positive difference between validation and testing performances was obtained for the Danish subset, which may indicate that the XLM-Roberta model could have been more suitable for the Danish language too, as the small difference obtained in the validation phase could have been outweighed by the generalizing  power given by the large multilingual fine-tuning dataset.

Moreover, the performances of the multilingual models increased not only with the size of the fine-tuning dataset, but also with the number of languages it contains.
The results also demonstrated that the potential of the multilingual Transformer-based models in offensive language detection could be improved if larger datasets are available for non-English languages. 
For future work, we intend to consider a transfer learning method in order to leverage datasets that  were constructed for similar tasks in the same language.

\bibliographystyle{coling}
\bibliography{coling2020}

\end{document}